\documentclass{article}

\usepackage{PRIMEarxiv}
\usepackage{amsmath,amssymb,amsfonts}
\usepackage{algorithmic}
\usepackage{xcolor}
\usepackage{multirow}
\usepackage{bm}
\usepackage[utf8]{inputenc} 
\usepackage[T1]{fontenc}    
\usepackage{hyperref}       
\usepackage{url}            
\usepackage{booktabs}       
\usepackage{amsfonts}       
\usepackage{nicefrac}       
\usepackage{microtype}      
\usepackage{lipsum}
\usepackage{fancyhdr}       
\usepackage{graphicx}       
\graphicspath{{media/}}     

\title{Statistical Non-linear Reconstruction Loss for Image Anomaly Detection
\thanks{\textit{\underline{Preprint}}: Accepted at the 30th International Conference on Knowledge-Based and Intelligent Information \& Engineering Systems (KES 2026).} 
}

\author{
  Nguyen Minh Tri\thanks{Equal contribution.} \quad
  Hoang Khuong Duy\footnotemark[2] \quad
  Huynh Cong Viet Ngu\thanks{Corresponding author. \emph{E-mail address:} \texttt{nguhcv@fe.edu.vn}} \\
  AIC-Lab, FPT University \\
  Ho Chi Minh City, Vietnam \\
}

\begin{document}
\maketitle

\begin{abstract}
Reconstruction-based methods are a cornerstone of unsupervised image anomaly detection, but they remain vulnerable to \emph{outlier leakage}, where standard mean squared error (MSE) loss drives the model to faithfully reconstruct anomalous patterns. We propose a Non-linear Reconstruction Loss that applies a sigmoid-based squashing function to suppress high-magnitude features, preventing outliers from dominating optimization while preserving sensitivity to normal patterns. In addition, we introduce a statistical calibration scheme that selects the scaling factor $k$ from the confidence interval (CI) of the normal feature distribution, enabling data-driven control of the suppression strength. Our approach achieves competitive or superior anomaly detection performance compared to state-of-the-art methods, reaching 99.0\% Image-AUROC and 97.3\% Pixel-AUROC on MVTec-AD, and 95.3\% Image-AUROC and 99.0\% Pixel-AUROC on VisA. These results indicate that non-linear gradient suppression is an effective mechanism for mitigating outlier leakage and improving anomaly localization in unified industrial inspection settings. The implementation is available at \href{https://github.com/mintii13/Statistical-Non-linear-Reconstruction-Loss.git}{https://github.com/mintii13/Statistical-Non-linear-Reconstruction-Loss.git}.
\end{abstract}


\section{Introduction}
\label{sec:intro}
Image Anomaly Detection (IAD) is a core component of modern industrial intelligence, supporting automated quality control, defect inspection, and security monitoring~\cite{bergmann2019mvtec, liu2024deep, sultani2018real}. Traditional approaches predominantly rely on a one-model-per-category paradigm, which imposes significant operational burdens due to poor scalability and high storage requirements; maintaining hundreds of dedicated models for diverse product lines is often computationally prohibitive and impractical for edge deployment. This has motivated the shift toward the more practical yet challenging Unified Framework setting~\cite{you2022unified}, where a single model must capture the diverse normal manifolds of multiple object categories simultaneously, offering a practical and resource-efficient solution for modern smart factories. 

Reconstruction-based methods, often built on autoencoders or Transformer architectures, are especially popular: a model trained only on normal data is expected to reconstruct anomalies poorly. However, reconstruction-based methods are frequently plagued by the identical shortcut phenomenon—a failure mode where the model settles for a trivial identity mapping rather than learning the underlying manifold of normality. This vulnerability is significantly exacerbated in the unified multi-class setting; as the model's capacity must scale to accommodate the structural diversity of multiple categories, it becomes increasingly prone to "copy-pasting" any input pattern, including anomalous ones, thereby losing its discriminative power.

\begin{figure}[h]
\centering
\includegraphics[width=0.78\linewidth]{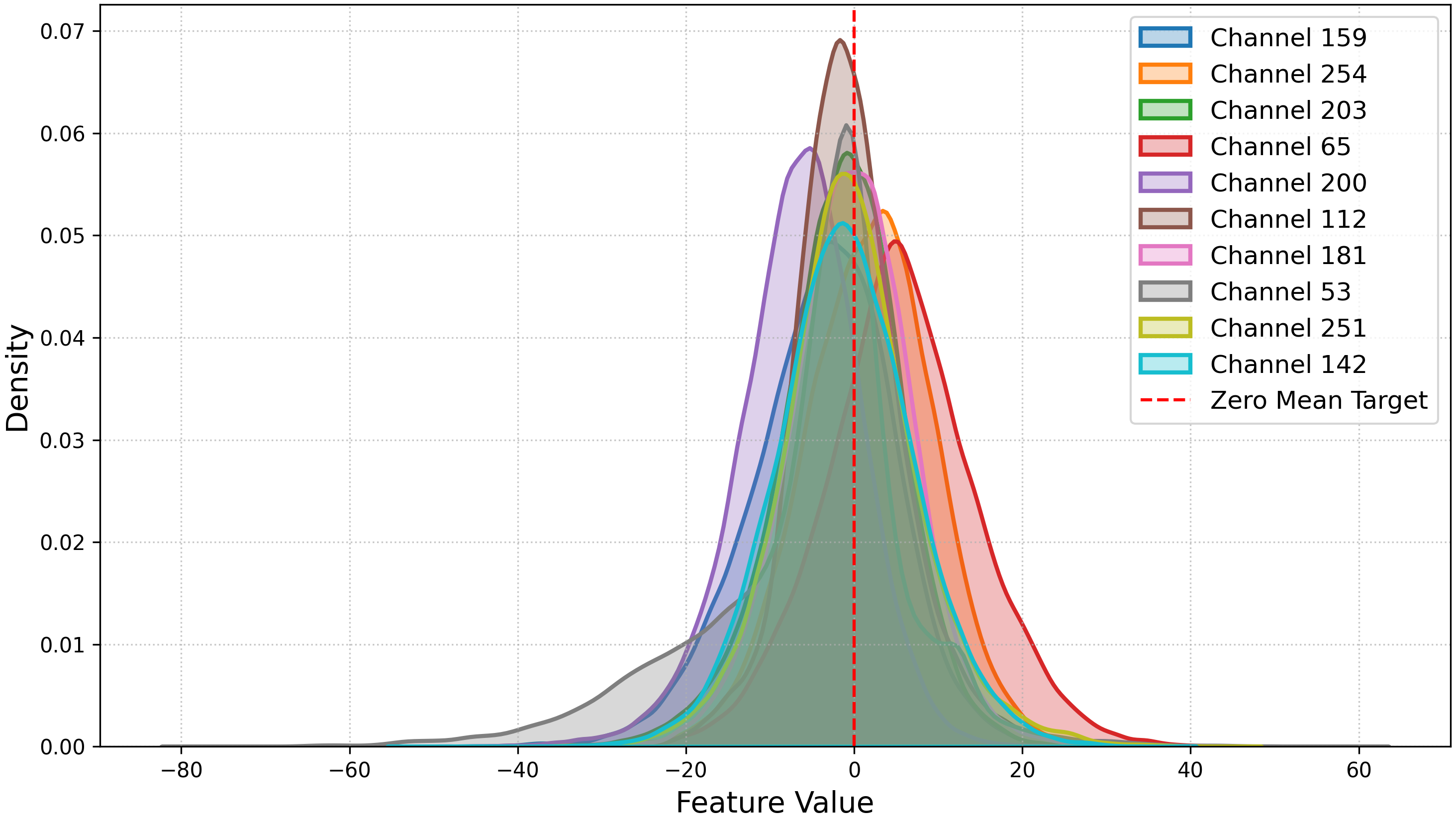}
\caption{Statistical density plot of 10 random channels of feature map after extracted by EfficientNet-B4 (Pretrained on ImageNet) \cite{you2022adtr}.}
\label{fig:feature_density}
\end{figure}

A prominent strategy to mitigate this is reconstructing features instead of raw pixels~\cite{you2022adtr}, leveraging high-level semantic representations from pretrained backbones to discourage trivial pixel-wise copying. As shown in Fig.~\ref{fig:feature_density}, we observed that extracted features are concentrated near zero, but contain high-magnitude outliers. These outliers represent potential anomalies in the data illustrated in Fig.~\ref{fig:nonlinear_logic}(a). Most existing reconstruction-based methods~\cite{hinton1994autoencoders, akcay2018ganomaly, gong2019memorizing, zavrtanik2021draem, you2022unified, you2022adtr, he2024diffusion, liang2021feature, shi2021unsupervised} rely on a standard Mean Squared Error (MSE) loss. As shown in Fig.~\ref{fig:nonlinear_logic}(b), this objective tries to reconstruct every feature equally—including these outliers—as accurately as possible. This forces the network to learn an 'identity-like' shortcut, where it simply copies the input features to the output. Because the model becomes too good at reconstructing even the abnormal parts, the gap between normal and abnormal data disappears, leading to a phenomenon we define as Outlier Leakage.



To address this issue, we propose a \textbf{Non-linear Reconstruction Loss} that squashes features through a non-linear activation before computing the reconstruction error. As shown in Fig.~\ref{fig:nonlinear_logic}(c), high-magnitude outliers fall into suppressed regions where gradients vanish, effectively masking them from optimization and focusing learning on the zero-centered normal distribution. We further introduce a \textbf{Statistical $k$-Value Calibration} scheme illustrated in Fig.~\ref{fig:nonlinear_logic}(d) that adapts the non-linearity by selecting the slope factor $k$ from the confidence interval (CI) of the training feature distribution, thereby aligning the suitable suppressed range or  normal region for different datasets. Our main contributions are:
\begin{itemize}
\item We introduce a Non-linear Reconstruction Loss to mitigate Outlier Leakage by suppressing the influence of high-magnitude potential anomalies during the optimization process.
\item We design a unified architecture that integrates the feature representations of a pretrained backbone with a Transformer-based autoencoder.
\item We demonstrate state-of-the-art anomaly detection performance on the MVTec-AD and VisA benchmarks, proving that our method significantly enhances detection performance by concentrating model capacity on the essential normal manifold.
\end{itemize}

The remainder of the paper is organized as follows: Section~\ref{sec:related} reviews related work in anomaly detection, Section~\ref{sec:method} describes the proposed methodology, Section~\ref{sec:exp} presents experimental results, and Section~\ref{sec:conclusion} concludes and outlines future directions.

\section{Related Work}
\label{sec:related}

\begin{figure}[h]
\centering
\includegraphics[width=0.9\linewidth]{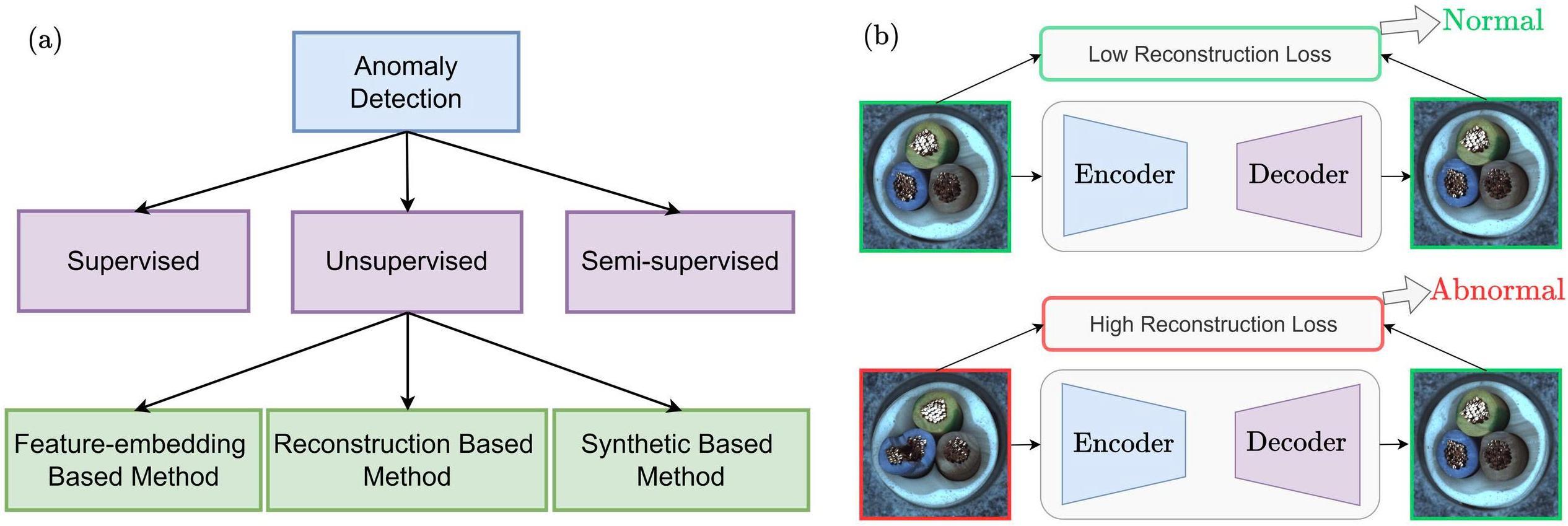}
\caption{(a) Taxonomy of methods based on supervision and approach, and (b) The fundamental logic of reconstruction-based anomaly detection.}
\label{fig:taxonomy}
\end{figure}

The field of Image Anomaly Detection (IAD) has evolved from classical statistical modeling to deep representation learning~\cite{liu2024deep, fernando2021deep}. As summarized in the taxonomy in Fig.~\ref{fig:taxonomy}(a), existing methods are typically grouped into supervised, semi-supervised, and unsupervised approaches, depending on the availability of anomaly labels. Because anomalous samples are scarce and often impractical to label in industrial settings, unsupervised methods that learn exclusively from normal data are generally preferred. Within this regime, most techniques fall into three categories: feature-embedding, reconstruction-based, and synthetic anomaly--based methods.

\subsection{Feature-Embedding Approaches}
These methods detect anomalies by comparing test images against a learned distribution of normal features. Teacher--student distillation models such as US~\cite{bergmann2020uninformed} and MKD~\cite{salehi2021multiresolution} measure discrepancies between a fixed teacher and a trainable student. Distribution-based methods, including PaDiM~\cite{defard2021padim} and PatchCore~\cite{roth2022towards}, explicitly model the local feature distribution. One-class techniques such as Patch SVDD~\cite{yi2020patch} and CutPaste~\cite{li2021cutpaste} learn a decision boundary around normal data. While effective, many of these approaches rely on large memory banks or struggle in unified multi-class settings. Recent work leverages Transformer-based feature modeling (e.g., ViTAD~\cite{zhang2025exploring}) and contrastive learning (e.g., CCL~\cite{fan2025salvaging}) to strengthen semantics and improve performance in unified scenarios.

\subsection{Reconstruction-Based Paradigms}
Reconstruction-based IAD as shown in Fig.~\ref{fig:taxonomy}(b) assumes that models trained only on normal data will fail to accurately reconstruct anomalies~\cite{hinton1994autoencoders, gong2019memorizing}. Autoencoder and GAN variants such as GANomaly~\cite{akcay2018ganomaly} and OCGAN~\cite{perera2019ocgan} implement this principle by enforcing high reconstruction error on abnormal regions. However, a fundamental challenge in this paradigm is the identical shortcut—a failure mode where the model learns a trivial identity mapping, allowing it to reconstruct any input (including anomalies) with high fidelity.

To address the "identical shortcut" problem, several advanced methods have been proposed. Transformer-based models like UniAD~\cite{you2022unified} and HVQ-Trans~\cite{lu2023hierarchical} utilize layer-wise query embeddings and vector quantization to restrict the model's reconstructive capability. However, HVQ-Trans relies on a multi-scale codebook mechanism, which leads to a linear increase in memory overhead as the number of categories grows—making it impractical for large-scale unified deployment. Meanwhile, although UniAD offers a more consolidated architecture, it still struggles with limited detection performance in complex industrial scenarios. More recent state-of-the-art models, such as DiAD~\cite{he2024diffusion} and MambaAD~\cite{he2024mambaad}, attempt to bypass the identical shortcut using generative diffusion processes or Selective State Space Modeling. Nevertheless, both DiAD and MambaAD still exhibit relatively low anomaly detection accuracy in unified settings. All these paradigms fundamentally rely on linear objectives (L1/L2) or structural similarity metrics~\cite{wang2004image, bergmann2018improving}, which remain vulnerable to Outlier Leakage when high-magnitude potential anomalies dominate the loss during training.


\subsection{Synthetic Anomaly--Based Methods}
\label{subsection:synthetic}
Another line of work augments training with synthetic anomalies to increase discriminability. DRAEM~\cite{zavrtanik2021draem} learns to distinguish normal images from artificially corrupted counterparts using a discriminatively trained reconstruction branch, while OneNIPs~\cite{gao2024learning} refines the normality boundary via advanced negative sampling. These methods can significantly improve detection quality but often depend on the realism and diversity of the generated anomalies.

\subsection{Robustness of Loss Objectives}
The sensitivity of mean squared error to outliers is a long-standing issue in regression~\cite{huber1992robust}. In the context of IAD, only a few works explicitly modify the reconstruction objective to address feature-level noise; structural similarity--based losses~\cite{bergmann2018improving} still penalize large discrepancies strongly and can be dominated by high-magnitude outliers. In contrast, our approach introduces a non-linear activation layer in the loss computation that leverages the feature distribution to saturate gradients for outliers, ensuring that optimization focuses on the core normal manifold rather than on rare, high-magnitude deviations.

\section{Methodology}
\label{sec:method}

\subsection{Overview of Architecture}
\label{sec:overview}

\begin{figure*}[h!]
\centering
\includegraphics[width=\linewidth]{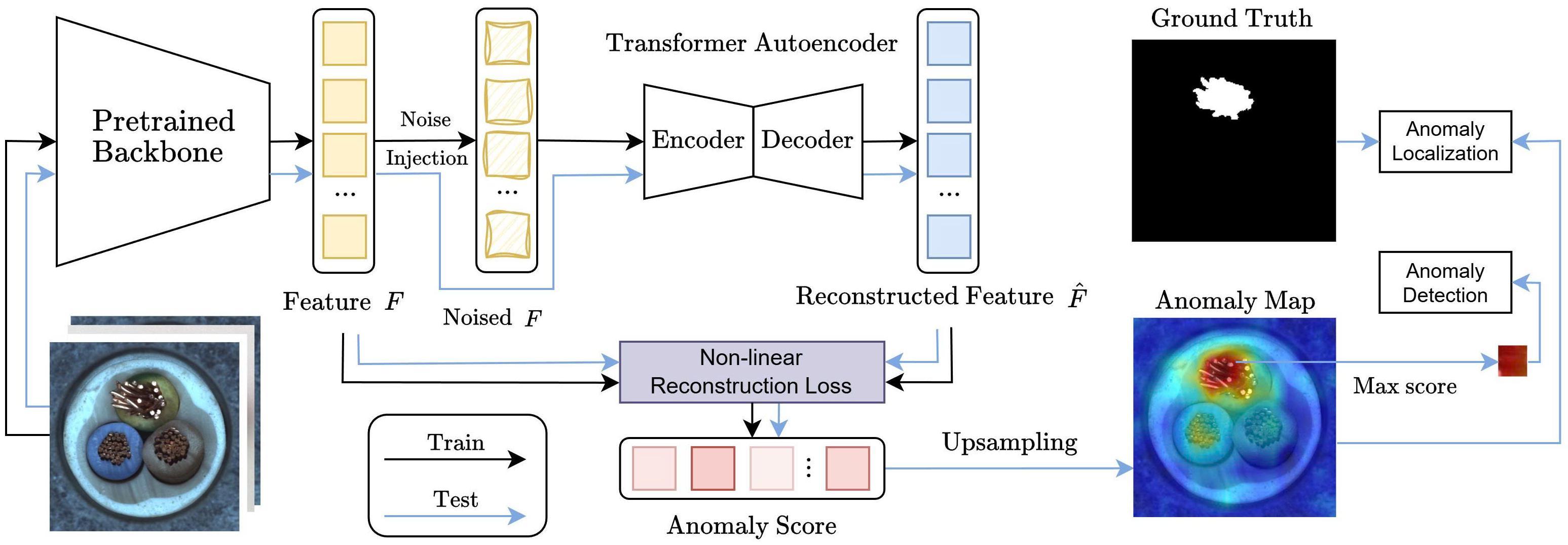}
\caption{The proposed anomaly detection pipeline. Features from the backbone are aligned and passed through a Transformer Autoencoder. The reconstruction loss is calculated in a non-linear space after statistical calibration, then upsampling for further anomaly detection and localization in Testing phase.}
\label{fig:overview}
\end{figure*}

The proposed network, illustrated in Fig.~\ref{fig:overview}, follows a reconstruction-based design~\cite{you2022adtr, shi2021unsupervised} comprising a pretrained backbone and a Transformer autoencoder operating in feature space. A pretrained backbone first extracts hierarchical representations from the input image; these multi-scale features are spatially aligned and concatenated into a unified tensor $\bm{F}$ that aggregates semantic information across layers. To encourage a more generalized notion of normality~\cite{bengio2013generalized, vincent2008extracting, you2022unified}, noise is added to the aligned features and the autoencoder is trained to reconstruct the corresponding clean features.

The vanilla Transformer autoencoder then processes $\bm{F}$ as a sequence of tokens and produces a reconstructed feature map $\hat{\bm{F}}$. Both $\bm{F}$ and $\hat{\bm{F}}$ are passed to the proposed Non-linear Reconstruction Loss, which computes an element-wise discrepancy and yields a feature-space anomaly score map. This map is subsequently upsampled to the original image resolution. For image-level anomaly detection, we take the maximum value of the upsampled anomaly map, reflecting the intuition that a single anomalous pixel can render an image abnormal. For anomaly localization, the same map provides pixel-level cues indicating regions that deviate from the learned normal manifold.

\subsection{Non-linear Reconstruction Loss}
\label{subsec:nonlinear_loss}

\begin{figure}[h!]
\centering
\includegraphics[width=1\linewidth]{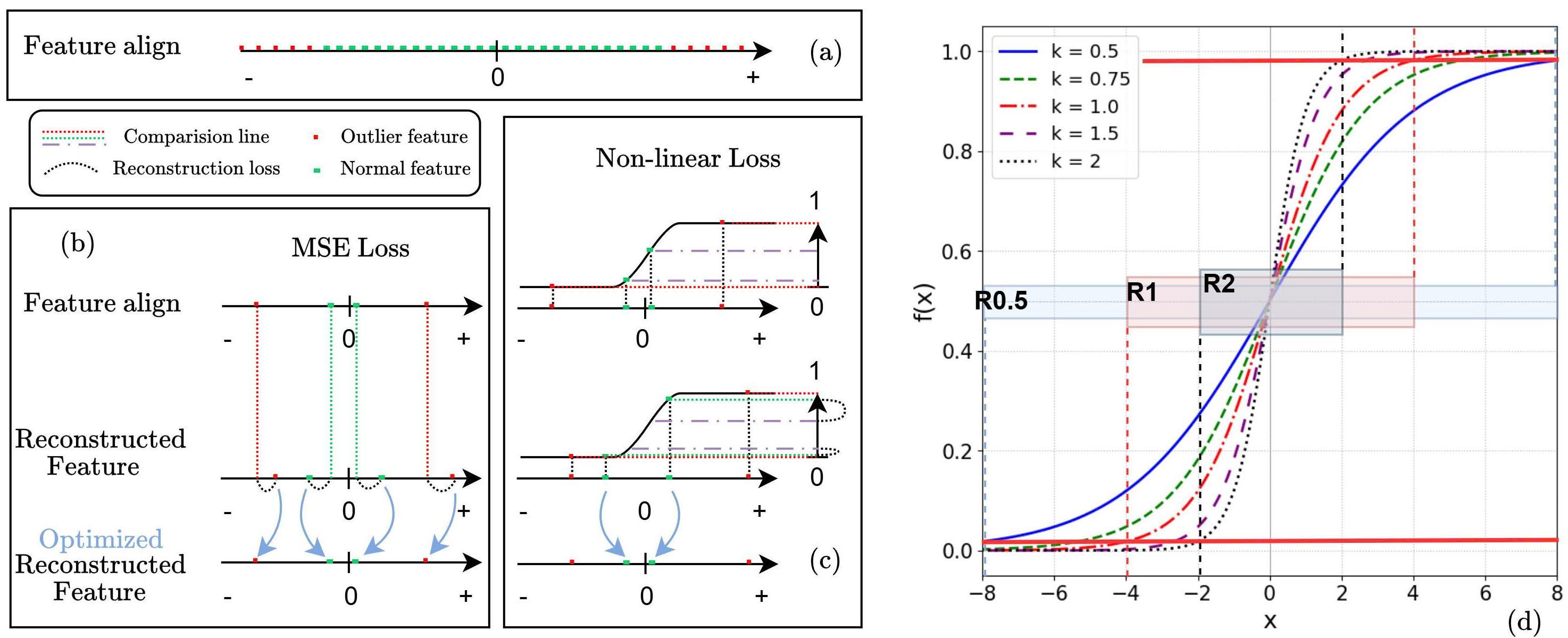}
\caption{Illustration of the proposed loss mechanism: (a) extracted features, (b) MSE loss optimization, (c) proposed Non-linear Loss with gradient suppression, and (d) dynamic adjustment of the active learning zones ($R$) across different slope factors $k$ derived from statistical calibration.}
\label{fig:nonlinear_logic}
\end{figure}

As observed in Section~\ref{sec:intro}, normal feature representations, after the pretrained backbone and alignment neck, lie in a band around zero, whereas noisy or anomalous features appear as high-magnitude outliers in the distribution as illustrated in Fig.~\ref{fig:nonlinear_logic}(a). Standard reconstruction-based anomaly detection typically adopts mean squared error (MSE):
\begin{equation}
\mathcal{L}_{\text{MSE}} = \lVert \bm{F} - \hat{\bm{F}} \rVert^{2},
\end{equation}
whose gradient with respect to the reconstructed feature $\hat{\bm{F}}$ is:
\begin{equation}
\frac{\partial \mathcal{L}_{\text{MSE}}}{\partial \hat{\bm{F}}}
= -2(\bm{F} - \hat{\bm{F}}).
\end{equation}

As the discrepancy $\lVert \bm{F} - \hat{\bm{F}} \rVert$ increases, the gradient grows linearly, forcing the model to devote its learning capacity to reconstructing all features—including high-magnitude outliers—as accurately as possible. This global optimization pressure leads to Outlier Leakage, where the network learns to reconstruct both normal patterns and potential anomalies with high fidelity, ultimately narrowing the discriminative gap between them. To mitigate this effect, we introduce a Non-linear Reconstruction Loss that squashes features into a bounded interval $(0,1)$ using a sigmoid activation $\sigma(\cdot)$:
\begin{equation}
\mathcal{L}_{\text{NL}}
= \big\lVert \sigma(k \bm{F}) - \sigma(k \hat{\bm{F}}) \big\rVert^{2},
\end{equation}
where $k$ is a slope factor determined by the statistical calibration in Section~\ref{subsectiob: stat}. The gradient with respect to $\hat{\bm{F}}$ is
\begin{equation}
\frac{\partial \mathcal{L}_{\text{NL}}}{\partial \hat{\bm{F}}}
= -2k \big[\sigma(k \bm{F}) - \sigma(k \hat{\bm{F}})\big]
      \,\sigma(k \hat{\bm{F}})\big(1 - \sigma(k \hat{\bm{F}})\big).
\end{equation}

The term $\sigma(k \hat{\bm{F}})(1 - \sigma(k \hat{\bm{F}}))$ implements a gradient suppression mechanism. For normal features with $k \bm{F} \approx 0$, we have $\sigma'(0) = 0.25$, yielding a stable, non-vanishing gradient that encourages precise reconstruction. As the term $k\bm{F}$ deviates from the origin and moves toward $\pm \infty$, the features transition into the extreme regions of the sigmoid curve where the function becomes nearly constant. This behavior is clearly illustrated in Fig.~\ref{fig:nonlinear_logic}(c): while normal features (green) land on the steep, high-gradient slope near zero, the high-magnitude outliers (red) are pushed toward the upper and lower plateaus of the curve. In these suppression zones, the sigmoid derivative $\sigma(k \hat{\bm{F}})(1 - \sigma(k \hat{\bm{F}}))$ drops toward zero, acting as a non-linear gate. As also shown in the mapping process of Fig.~\ref{fig:nonlinear_logic}(c), even if there is a large discrepancy between the input and the reconstructed feature in the original space, this difference is "squashed" in the non-linear space. Consequently, the gradient is suppressed at both ends of the distribution, effectively preventing potential anomalies from influencing the backpropagation process. This ensures that the model ignores these outlying deviations and remains strictly focused on learning the core normal manifold. Crucially, this gradient suppression does not require the features to literally reach $\pm \infty$. Due to the rapid exponential decay of the sigmoid derivative, the gradient becomes negligible as soon as $|k\bm{F}|$ moves a sufficient distance from the origin (e.g., dropping below $0.019$ for $|k\bm{F}| > 4$). To efficiently determine the normal range or suppression zone, we propose the Statistical k-Value Calibration in Section~\ref{subsectiob: stat} to control the slope of non-linear activation.

\subsection{Statistical $k$-Value Calibration}
\label{subsectiob: stat}
To ensure that the sigmoid activation effectively isolates normal features from high-magnitude outliers, we propose a data-driven calibration for the slope factor $k$ based on the statistical distribution of the whole training set before training process. The global feature distribution often contains extreme values, yet the core normal patterns reside within a significantly tighter interval. By selecting a specific Confidence Interval ($CI$), we define the representative normal range $r$ of the extracted features and derive the slope factor $k$ as:
\begin{equation}
r = P_{\alpha_{high}} - P_{\alpha_{low}}, \qquad k = \frac{\eta}{r}
\end{equation}
where $P_{\alpha}$ denotes the $\alpha$-th percentile of the aligned feature distribution. To center the interval on the distribution's core, the upper and lower percentile bounds are defined as $\alpha_{high} = \frac{100+CI}{2}$ and $\alpha_{low} = \frac{100-CI}{2}$, respectively (e.g., for $CI = 90\%$, $r = P_{95} - P_{5}$). In our formulation, $\eta$ serves as a fixed scaling constant representing the "standardized" high-gradient window of the sigmoid curve.  Mathematically, $k$ acts as a scaling operator that adjusts the steepness of the sigmoid curve to match the density of the data:

\begin{itemize}\item If the normal features are widely spread (large $r$), a smaller $k$ is calculated to broaden the active learning zone.\item If the features are tightly clustered (small $r$), a larger $k$ is derived to sharpen the curve's sensitivity.\end{itemize}As visualized in Fig.~\ref{fig:nonlinear_logic}(d), $k$ directly controls the width of the "normal feature regions" ($R$), where the sigmoid derivative remains significant. For instance, as $k$ increases from 0.5 to 2.0, the active learning zone $R$ compresses from a broad range $[-8, 8]$ down to a narrow window of $[-2, 2]$. To illustrate this, consider a scenario where raw features range from $-100$ to $100$, but the $90\%$ CI identifies the normal manifold within the interval $[-8, 8]$, resulting in $r = 16$. By applying our formulation, we derive $k = \eta / 16 $. During training, this calibrated $k$ ensures that normal features—which fluctuate near the zero mean—are mapped into the high-sensitivity region of the sigmoid. Conversely, potential anomalies extending far beyond the statistical CI are automatically pushed into the "suppression zones" where the derivative approaches zero. Consequently, these outliers produce near-zero gradients, effectively masking them from the backpropagation process and preventing Outlier Leakage by focusing the model's capacity exclusively on the statistically defined normal manifold.

Regarding hyperparameter complexity, it is important to clarify that $CI$ and $k$ do not function as two independent variables requiring simultaneous manual tuning. Instead, $CI$ represents our statistical assumption of the normal manifold's boundary, while $k$ is the corresponding slope for the non-linear activation, automatically derived based on the training data's distribution. The calibration of $k$ is conducted exclusively on the entire training set prior to inference. This protocol ensures that the non-linear suppression range is strictly informed by the normal distribution, completely preventing any data leakage from the test set. In our unified framework, a single calibrated $k$ generalizes effectively across all object and texture categories within a given dataset. However, because $k$ is intrinsically linked to the underlying feature statistics, its value naturally adapts across different benchmarks (e.g., MVTec-AD vs. VisA) to accommodate their distinct feature densities. The optimization and sensitivity analysis of the Confidence Interval are further detailed in Section~\ref{sec:ablation}.

\section{Experiments}
\label{sec:exp}

\subsection{Experimental Setup}
\textbf{Datasets:} We evaluate our framework on two widely recognized industrial benchmarks: MVTec-AD~\cite{bergmann2019mvtec} and VisA.~\cite{zou2022spot} The MVTec-AD dataset consists of 15 categories, including 10 object classes and 5 texture classes, encompassing various real-world defect scenarios. The VisA dataset provides a more rigorous challenge with 12 categories organized into three subsets: Complex Structures, Multiple Instances, and Single Instances.

\textbf{Configuration:} The implementation uses PyTorch and is trained on the Kaggle platform utilizing an NVIDIA Tesla P100 GPU. All input images are resized to a resolution of $256 \times 256$ pixels. We employ the AdamW optimizer with a base learning rate of $1 \times 10^{-4}$ and a batch size of 8. For the training schedule, MVTec-AD is trained for 500 epochs, while the VisA dataset is trained for 250 epochs. To ensure stable convergence, a step learning rate scheduler is applied to reduce the learning rate at 80\% of the total training epochs. For reproducibility, the random seed is fixed at 42 across all experiments. We set $\eta = 8.0$ in order to map the raw range $r$ onto the standardized interval $[-4, 4]$ in the non-linear space to match with the observation at the end of Section~\ref{subsec:nonlinear_loss}.

\textbf{SOTA Comparisons:} Our proposed method is compared against a comprehensive suite of state-of-the-art models, including UniAD~\cite{you2022unified}, HVQ-Trans~\cite{lu2023hierarchical}, DiAD~\cite{he2024diffusion}, OneNIPs~\cite{gao2024learning}, CCL~\cite{fan2025salvaging}, ViTAD~\cite{zhang2025exploring}, and MambaAD~\cite{he2024mambaad}. \textit{Note that while CCL~\cite{fan2025salvaging} utilizes class labels during training, which introduces additional complexity to the standard unified setting as~\cite{you2022unified} has mentioned, we report and compare against their results obtained without the use of class labels to ensure a fair and consistent evaluation.}

\textbf{Metrics:} To provide a thorough quantitative assessment, we utilize two standard industrial metrics. Image-level AUROC (Image-AUROC) is employed as the primary metric for anomaly detection to evaluate the model's ability to classify an entire image as normal or abnormal. Pixel-level AUROC (Pixel-AUROC) is used for anomaly localization to measure the precision with which the model identifies and segments specific anomalous regions within an image.

\begin{table*}[h]
\centering
\caption{Unified anomaly detection (Image-level) and localization (Pixel-level) performance compared to state-of-the-art models on MVTec-AD.}
\label{tab:mvtec}
\resizebox{\linewidth}{!}{%
\begin{tabular}{cccccccccccccccccc}
\hline
\multicolumn{2}{c}{\multirow{3}{*}{\textbf{Class}}} & \multicolumn{2}{c}{\textbf{UniAD}} & \multicolumn{2}{c}{\textbf{HVQ-Trans}} & \multicolumn{2}{c}{\textbf{DiAD}} & \multicolumn{2}{c}{\textbf{OneNIPs}} & \multicolumn{2}{c}{\textbf{CCL}} & \multicolumn{2}{c}{\textbf{ViTAD}} & \multicolumn{2}{c}{\textbf{MambaAD}} & \multicolumn{2}{c}{\textbf{Ours}} \\
\multicolumn{2}{c}{} & \multicolumn{2}{c}{\scriptsize NeurIPS'22} & \multicolumn{2}{c}{\scriptsize NeurIPS'23} & \multicolumn{2}{c}{\scriptsize AAAI'24} & \multicolumn{2}{c}{\scriptsize ECCV'24} & \multicolumn{2}{c}{\scriptsize ICCV'25} & \multicolumn{2}{c}{\scriptsize CVIU Q1'25} & \multicolumn{2}{c}{\scriptsize NeurIPS'25} & \multicolumn{2}{c}{-} \\
\multicolumn{2}{c}{} & Image & Pixel & Image & Pixel & Image & Pixel & Image & Pixel & Image & Pixel & Image & Pixel & Image & Pixel & Image & Pixel \\ \hline
\multirow{10}{*}{\textbf{Object}} & bottle & 99.7 & 98.1 & \textbf{100} & 98.3 & 99.7 & 98.4 & \textbf{100} & 98.5 & 97.7 & \textbf{99.6} & \textbf{100} & 98.8 & \textbf{100} & 98.8 & \textbf{100} & 97.8 \\
 & cable & 95.2 & 97.3 & 99.0 & 98.1 & 94.8 & 96.8 & \textbf{99.0} & \textbf{98.2} & 82.3 & 94.7 & 98.5 & 96.2 & 98.8 & 95.8 & 97.8 & 96.3 \\
 & capsule & 86.9 & 98.5 & 95.4 & 98.8 & 89.0 & 97.1 & 91.1 & 98.6 & 97.6 & 85.3 & 95.4 & 98.3 & 94.4 & 98.4 & \textbf{98.5} & \textbf{98.8} \\
 & hazelnut & 99.8 & 98.1 & \textbf{100} & 98.8 & 99.5 & 98.3 & \textbf{100} & 98.7 & 99.0 & \textbf{100} & 99.8 & 99.0 & \textbf{100} & 99.0 & \textbf{100} & 98.2 \\
 & metalnut & 99.2 & 94.8 & 99.9 & 96.3 & 99.1 & 97.3 & 99.8 & 96.5 & 95.8 & \textbf{99.7} & 99.7 & 96.4 & 99.9 & 96.7 & \textbf{100} & 95.2 \\
 & pill & 93.7 & 95.0 & 95.8 & 97.1 & 95.7 & 95.7 & 96.9 & 96.0 & 96.9 & 91.8 & 96.2 & \textbf{98.7} & 97.0 & 97.4 & \textbf{98.7} & 97.0 \\
 & screw & 87.5 & 98.3 & 95.6 & 98.9 & 90.7 & 97.9 & 91.4 & 98.9 & 98.6 & 86.7 & 91.3 & 99.0 & 94.7 & 99.5 & \textbf{99.0} & \textbf{99.6} \\
 & toothbrush & 94.2 & 98.4 & 93.6 & 98.6 & 99.7 & 99.0 & 93.3 & 98.8 & 85.5 & 94.1 & \textbf{98.9} & \textbf{99.1} & 98.3 & 99.0 & 93.3 & 98.3 \\
 & transistor & 99.8 & 97.9 & 99.7 & 97.9 & 99.7 & 95.1 & 99.8 & 98.8 & 99.0 & \textbf{99.4} & 98.8 & 93.9 & \textbf{100} & 96.5 & 99.7 & 96.2 \\
 & zipper & 95.8 & 96.8 & 97.9 & 97.5 & 95.1 & 96.2 & 99.0 & 97.6 & 97.2 & \textbf{99.4} & 97.6 & 95.9 & 99.3 & 98.4 & \textbf{99.9} & 98.5 \\ \hline
\multirow{5}{*}{\textbf{Texture}} & carpet & 99.8 & 98.5 & 99.9 & 98.7 & 99.4 & 98.6 & \textbf{99.9} & 99.0 & 99.0 & \textbf{99.2} & 99.5 & 99.0 & 99.8 & \textbf{99.2} & \textbf{99.9} & 98.6 \\
 & grid & 98.2 & 96.5 & 97.0 & 97.0 & 98.5 & 96.6 & 99.0 & 98.4 & 98.3 & 98.7 & 99.7 & 98.6 & \textbf{100} & \textbf{99.2} & \textbf{100} & 98.0 \\
 & leather & \textbf{100} & 98.8 & \textbf{100} & 98.8 & 99.8 & 98.8 & \textbf{100} & 99.6 & 99.4 & \textbf{100} & \textbf{100} & 99.6 & \textbf{100} & 99.4 & \textbf{100} & 99.3 \\
 & tile & 99.3 & 91.8 & 99.2 & 92.2 & 96.8 & 92.4 & \textbf{100} & 95.3 & 95.3 & \textbf{99.6} & \textbf{100} & 96.6 & 98.2 & 93.8 & 99.8 & 94.1 \\
 & wood & 98.6 & 93.2 & 97.2 & 92.4 & \textbf{99.7} & 93.3 & 98.8 & 94.9 & 95.3 & \textbf{99.0} & 98.7 & 96.4 & 98.8 & 94.4 & 99.0 & 94.2 \\ \hline
\multicolumn{2}{c}{\textbf{Mean}} & 96.5 & 96.8 & 98.0 & 97.3 & 97.2 & 96.8 & 97.9 & \textbf{97.9} & 96.5 & 95.8 & 98.3 & 97.7 & 98.6 & 97.7 & \textbf{99.0} & 97.3 \\ \hline
\end{tabular}%
}
\end{table*}

\subsection{Results on MVTec-AD}

As presented in Table~\ref{tab:mvtec}, our method achieves a leading mean Image-level AUROC of \textbf{99.0\%}, outperforming strong baselines such as MambaAD (98.6\%) and ViTAD (98.3\%). Notably, our model reaches a perfect 100\% Image AUROC in several categories including bottle, hazelnut, metalnut, grid, and leather. In terms of anomaly localization, we achieve a competitive mean Pixel-level AUROC of \textbf{97.3\%}, with state-of-the-art performance in specific classes like capsule (98.8\%) and screw (99.6\%). While our mean localization score is slightly lower than OneNIPs (97.9\%) on MVTec-AD, this discrepancy can be attributed to the nature of OneNIPs as a synthetic anomaly-based method. As discussed in Section~\ref{subsection:synthetic}, the effectiveness of such paradigms is heavily dependent on the diversity and realism of the artificially generated anomalies. While synthetic methods can excel when training defects closely resemble the test set, their generalization is often limited; this is evidenced by the fact that OneNIPs falls behind our method on the VisA dataset in both Image-level (92.5\% vs. 95.3\%) and Pixel-level (98.7\% vs. 99.0\%) metrics. These results confirm the superior robustness of our reconstruction-based approach in suppressing outlier leakage across diverse categories without relying on dataset-specific synthetic generation.

\subsection{Results on VisA}

As shown in Table~\ref{tab:visa}, our method demonstrates superior performance, achieving a state-of-the-art mean Image-level AUROC of \textbf{95.3\%}. Notably, we achieve a mean Pixel-level AUROC of \textbf{99.0\%}, confirming that the non-linear loss mechanism provides unmatched precision for anomaly localization in complex scenarios. This performance is highly consistent across challenging categories, including complex structures (PCB) and multiple instances (Capsules, Macaroni), where traditional MSE-based models often struggle with high-magnitude feature noise.

\begin{table*}[h]
\centering
\caption{Unified anomaly detection (Image-level) and localization (Pixel-level) performance compared to state-of-the-art models on VisA dataset.}
\label{tab:visa}
\resizebox{\linewidth}{!}{%
\begin{tabular}{cccccccccccccccc}
\hline
\multicolumn{2}{c}{\multirow{3}{*}{\textbf{Class}}} & \multicolumn{2}{c}{\textbf{HVQ-Trans}} & \multicolumn{2}{c}{\textbf{DiAD}} & \multicolumn{2}{c}{\textbf{OneNIPs}} & \multicolumn{2}{c}{\textbf{CCL}} & \multicolumn{2}{c}{\textbf{ViTAD}} & \multicolumn{2}{c}{\textbf{MambaAD}} & \multicolumn{2}{c}{\textbf{Ours}} \\
\multicolumn{2}{c}{} & \multicolumn{2}{c}{\scriptsize NeurIPS'23} & \multicolumn{2}{c}{\scriptsize AAAI'24} & \multicolumn{2}{c}{\scriptsize ECCV'24} & \multicolumn{2}{c}{\scriptsize ICCV'25} & \multicolumn{2}{c}{\scriptsize CVIU Q1'25} & \multicolumn{2}{c}{\scriptsize NeurIPS'25} & \multicolumn{2}{c}{-} \\
\multicolumn{2}{c}{} & Image & Pixel & Image & Pixel & Image & Pixel & Image & Pixel & Image & Pixel & Image & Pixel & Image & Pixel \\ \hline
\multirow{4}{*}{\textbf{Complex Str.}} & PCB 1 & 96.7 & 99.4 & 88.1 & 98.7 & 95.8 & 99.6 & 96.7 & 99.5 & 95.8 & 99.5 & 95.4 & \textbf{99.8} & \textbf{98.8} & 99.6 \\
 & PCB 2 & 93.4 & 98.0 & 91.4 & 95.2 & 94.1 & 98.1 & 97.8 & 98.0 & 90.6 & 97.9 & 94.2 & \textbf{98.9} & \textbf{98.2} & 98.8 \\
 & PCB 3 & 92.0 & 98.3 & 86.2 & 96.7 & 91.9 & 98.2 & 96.7 & 98.1 & 90.9 & 98.2 & 93.7 & \textbf{99.1} & \textbf{96.9} & 99.0 \\
 & PCB 4 & 99.5 & 97.7 & 99.6 & 97.0 & 99.5 & 98.1 & \textbf{100.0} & 97.8 & 99.1 & \textbf{99.1} & 99.9 & 98.6 & 99.4 & 98.7 \\ \hline
\multirow{4}{*}{\textbf{Multiple Inst.}} & Macaroni 1 & 93.1 & 99.4 & 85.7 & 94.1 & 91.9 & 99.2 & \textbf{95.8} & \textbf{99.8} & 85.8 & 98.5 & 91.6 & 99.5 & 92.4 & 99.6 \\
 & Macaron 2 & 86.2 & 98.5 & 62.5 & 93.6 & 84.1 & 97.9 & \textbf{87.7} & \textbf{99.6} & 79.1 & 98.1 & 81.6 & 99.5 & 85.4 & 99.2 \\
 & Capsules & 77.1 & 99.0 & 58.2 & 97.3 & 79.0 & 98.4 & 83.8 & \textbf{99.4} & 79.2 & 98.2 & \textbf{91.8} & 99.1 & 89.1 & 99.3 \\
 & Candles & 96.8 & 99.2 & 92.8 & 97.3 & 96.8 & 99.2 & 92.5 & 99.2 & 90.4 & 96.2 & 96.8 & 99.0 & \textbf{96.8} & \textbf{99.3} \\ \hline
\multirow{4}{*}{\textbf{Single Inst.}} & Cashew & \textbf{94.9} & \textbf{99.2} & 91.5 & 90.9 & 93.7 & \textbf{99.2} & 94.3 & 92.3 & 87.8 & 98.5 & 94.5 & 94.3 & 92.9 & 98.4 \\
 & Chewing gum & 99.4 & 98.8 & 99.1 & 94.7 & 99.3 & 99.1 & 96.4 & 98.8 & 94.9 & 97.8 & 97.7 & 98.1 & \textbf{99.6} & \textbf{99.3} \\
 & Fryum & 90.4 & \textbf{97.7} & 89.8 & 97.6 & 86.9 & \textbf{97.7} & \textbf{95.9} & 96.9 & 94.3 & 97.5 & 95.2 & 96.9 & 95.5 & 97.2 \\
 & Pipe fryum & 98.5 & 99.4 & 96.2 & 99.4 & 97.3 & \textbf{99.5} & 98.0 & 99.1 & 97.8 & \textbf{99.5} & 98.7 & 99.1 & \textbf{98.8} & 99.1 \\ \hline
\multicolumn{2}{c}{\textbf{Mean}} & 93.2 & 98.7 & 86.8 & 96.0 & 92.5 & 98.7 & 93.6 & 98.2 & 90.5 & 98.2 & 94.3 & 98.5 & \textbf{95.3} & \textbf{99.0} \\ \hline
\end{tabular}%
}
\end{table*}

\subsection{Ablation Study and Quantitative Results}
\label{sec:ablation}

\begin{figure}[h]
\centering
\includegraphics[width=0.96\linewidth]{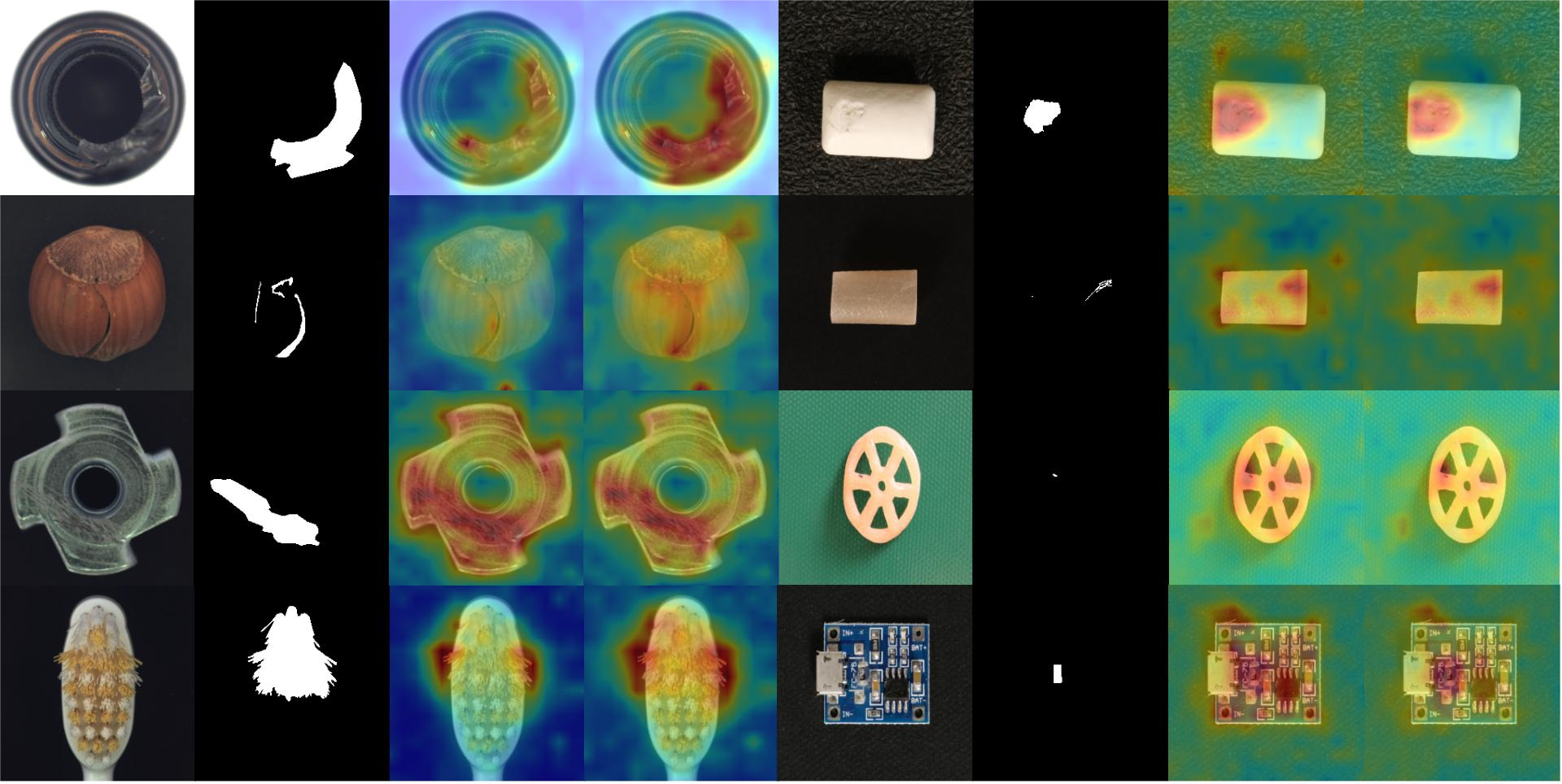}
\caption{Visual comparison of defect localization. The figure presents MVTec-AD results on the left and VisA results on the right. For each object sample, the columns from left to right represent: (a) Test Image, (b) Ground Truth mask, (c) Anomaly map from a baseline model, and (d) Ours.}
\label{fig:localization_visual_compare}
\end{figure}

Table~\ref{tab:ablation} demonstrates that our method achieves optimal performance at \textbf{CI 85\%} for MVTec-AD and \textbf{CI 95\%} for VisA, while significant drops in the absence of non-linear suppression (w/o NL) or calibration (w/o Cal.) confirm the necessity of our proposed modules. These findings validate that mitigating Outlier Leakage through data-driven $k$-value calibration substantially improves detection and localization efficiency. Furthermore, qualitative results in Fig.~\ref{fig:localization_visual_compare} show that our approach produces cleaner anomaly maps with sharper boundaries and reduced background noise compared to standard baselines, ensuring precise defect localization across both benchmarks.

\begin{table*}[h]
\centering
\caption{Ablation study on proposed modules and Calibration CI (Ordered Layout).}
\label{tab:ablation}
\begin{tabular}{llcccccccccc}
\toprule
\multirow{2}{*}{\textbf{Dataset}} & \multirow{2}{*}{\textbf{Metric}} & \multicolumn{2}{c}{\textbf{Baselines}} & \multicolumn{8}{c}{\textbf{Non-linear Loss with Confidence Interval (CI)}} \\
\cmidrule(lr){3-4} \cmidrule(lr){5-12}
& & \textbf{w/o NL} & \textbf{w/o Cal.} & \textbf{80} & \textbf{83} & \textbf{85} & \textbf{88} & \textbf{90} & \textbf{93} & \textbf{95} & \textbf{98} \\
\midrule
\multirow{2}{*}{\textbf{MVTec-AD}} & Image & 97.92 & 98.80 & 99.03 & 98.97 & \textbf{99.04} & 99.00 & 98.86 & 98.95 & 98.86 & 98.81 \\
& Pixel & 96.80 & 97.21 & 97.33 & 97.33 & \textbf{97.34} & 97.32 & 97.31 & 97.34 & 97.30 & 97.31 \\
\midrule
\multirow{2}{*}{\textbf{VisA}}     & Image & 94.42 & 95.00 & 95.23 & 95.17 & 95.09 & 95.16 & 95.19 & 95.14 & \textbf{95.33} & 95.17 \\
& Pixel & 98.79 & 98.00 & 98.92 & 98.94 & 98.96 & 98.96 & 98.94 & 98.96 & \textbf{98.96} & 98.96 \\
\bottomrule
\end{tabular}
\end{table*}

\section{Conclusion}
\label{sec:conclusion}
In this study, we introduced a novel Non-linear Reconstruction Loss designed to mitigate the ``outlier leakage'' problem in Reconstruction-based Image Anomaly Detection. By employing a sigmoid-based squashing function and a data-driven Statistical $k$-Value Calibration mechanism, our model effectively suppresses high-magnitude feature noise, ensuring that the model's optimization capacity remains focused on the core normal manifold. The effectiveness of this approach is validated by extensive experiments on two major benchmarks, where our method achieves comparable or superior Anomaly Detection performance compared to state-of-the-art methods. On MVTec-AD, we achieved a mean Image-level AUROC of 99.0\% and a Pixel-level AUROC of 97.3\%. Similarly, on the more complex VisA dataset, our model reached an Image-level AUROC of 95.3\% and a leading Pixel-level AUROC of 99.0\%. By effectively mitigating the outlier leakage and identical shortcut problems, this study provides a highly robust and precise solution for automated defect inspection, significantly enhancing the reliability and efficiency of real-world industrial quality assurance systems. Currently, the method requires selecting an optimal CI range based on the dataset's characteristics. In the future, we will explore advanced statistical mechanisms to automatically capture the optimal normal range.

\bibliographystyle{unsrt}  
\bibliography{references}

\end{document}